\documentclass[11pt,a4paper]{article}
\usepackage[T5,T1]{fontenc}
\usepackage[utf8]{inputenc}
\usepackage{linguex}
\usepackage{authblk}
\usepackage[hyperref]{naaclhlt2018}
\usepackage{times}
\usepackage{latexsym}
\usepackage{amssymb}
\usepackage{amsmath}
\usepackage{pifont}
\usepackage{tikz}
\usepackage{tikz-dependency}
\usepackage{bbm}
\usepackage{tabularx}
\usepackage{array}
\usepackage{booktabs}
\usepackage{multirow}
\newcolumntype{L}[1]{>{\raggedright\let\newline\\\arraybackslash\hspace{0pt}}m{#1}}
\newcolumntype{C}[1]{>{\centering\let\newline\\\arraybackslash\hspace{0pt}}m{#1}}
\newcolumntype{R}[1]{>{\raggedleft\let\newline\\\arraybackslash\hspace{0pt}}m{#1}}

\usepackage{url}

\aclfinalcopy 
\def\aclpaperid{	212} 

\DeclareTextSymbolDefault{\OHORN}{T5}
\DeclareTextSymbolDefault{\UHORN}{T5}
\DeclareTextSymbolDefault{\ohorn}{T5}
\DeclareTextSymbolDefault{\uhorn}{T5}
\title{Sentences with Gapping: Parsing and Reconstructing Elided Predicates}

\author[ ]{\bf Sebastian Schuster}
\author[$\ddag$]{\bf Joakim Nivre}
\author[ ]{\bf Christopher D. Manning}
\affil[ ]{Departments of Linguistics and Computer Science, Stanford University \protect\\
\texttt{\{sebschu,manning\}@stanford.edu}}
\affil[$\ddag$]{Department of Linguistics and Philology, Uppsala University \protect\\
\texttt{joakim.nivre@lingfil.uu.se}}

\date{}

\begin{document}
\maketitle
\begin{abstract}
Sentences with gapping, such as {\it Paul likes coffee and Mary tea}, lack an overt predicate to indicate the relation between two or more arguments. Surface syntax representations of such sentences are often produced poorly by parsers, and even if correct, not well suited to downstream natural language understanding tasks such as relation extraction that are typically designed to extract information from sentences with canonical clause structure. In this paper, we present two methods for parsing to a Universal Dependencies graph representation that explicitly encodes the elided material with additional nodes and edges. We find that both methods can reconstruct elided material from dependency trees with high accuracy when the parser correctly predicts the existence of a gap. We further demonstrate that one of our methods can be applied to other languages based on a case study on Swedish. 

  \end{abstract}

\section{Introduction}

Sentences with gapping \cite{Ross1970} such as {\it Paul likes coffee and Mary tea}
are characterized by having one or more conjuncts that contain multiple arguments or modifiers
of an elided predicate. In this example, the predicate {\it likes} is elided for the relation {\it Mary likes tea}.
While these sentences appear relatively infrequently in most 
written texts, they are often used to convey a lot of factual information that is highly relevant for
language understanding (NLU) tasks such as open information extraction and semantic parsing. For example, 
consider the following sentence from the WSJ portion of the Penn 
Treebank \cite{Marcus1993}.
\ex. 
Unemployment has reached 27.6\% in Azerbaijan, 
25.7\% in Tadzhikistan, 22.8\% in Uzbekistan, 
18.8\% in Turkmenia, 18\% in Armenia and 16.3\% in Kirgizia, [...]

To extract the information about unemployment rates in the various countries, an NLU system has to identify that the percentages indicate 
unemployment rates and the locational modifiers indicate the corresponding country.
Given only this sentence, or this sentence and a strict surface syntax representation 
that does not indicate elided predicates, this is a challenging task. However, 
given a dependency graph that reconstructs the elided predicate for each
conjunct, the problem becomes much easier and methods developed 
to extract information from dependency trees of clauses with canonical structures
are much more likely to extract the correct information from a gapped clause.

\begin{figure}
\center
{\footnotesize
Paul likes coffee and Mary tea}

\begin{tikzpicture}
\clip (-1,0) rectangle (1, 0.4);
\draw[thick,->] (-.5,0.3) -- (-1,0);
\draw[thick,->] (.5,0.3) -- (1,0);
 \end{tikzpicture}
 
\begin{dependency}
\scriptsize
\begin{deptext}
Paul \& likes \& coffee \& and \& Mary \& tea \& \ \ \ \ \ \ \ \& Paul \& likes \& coffee \& and \& Mary \& tea\\
\end{deptext}

\depedge[edge unit distance=1.85ex]{2}{1}{nsubj}
\depedge[edge unit distance=1.85ex]{2}{3}{obj}
\depedge[edge unit distance=2.2ex]{2}{4}{\bf conj$>$cc}
\depedge[edge unit distance=2.35ex]{2}5{\bf conj$>$nsubj}
\depedge[edge unit distance=2.4ex]{2}{6}{\bf conj$>$obj}

\depedge{9}{8}{nsubj}
\depedge{9}{10}{obj}
\depedge{12}{11}{cc}
\depedge[edge unit distance=3ex]{12}{9}{conj}
\depedge{12}{13}{\bf{orphan}}

\end{dependency}

\begin{tikzpicture}
\clip (-1,-0.15) rectangle (1, 0.4);
\draw[thick,->] (-1,0.3) -- (-.5,0);
\draw[thick,->] (1,0.3) -- (.5,0);
 \end{tikzpicture}

\begin{dependency}
\footnotesize
\begin{deptext}
Paul \& likes \& coffee \& and \& Mary \& {\bf likes$'$} \& tea \\
\end{deptext}
\depedge[edge unit distance=2ex]{2}{1}{nsubj}
\depedge[edge unit distance=2ex]{2}{3}{obj}
\depedge[edge unit distance=2ex]{6}{4}{cc}
\depedge[edge unit distance=2ex]{6}{5}{nsubj}
\depedge[edge unit distance=1.35ex]{2}{6}{conj}
\depedge[edge unit distance=2ex]{6}{7}{obj}
\end{dependency}
\caption{\label{fig:overview} Overview of our two approaches. Both methods first parse a sentence with gapping to one of two different dependency tree representations and then
 reconstruct the elided predicate from this tree.}
\end{figure}
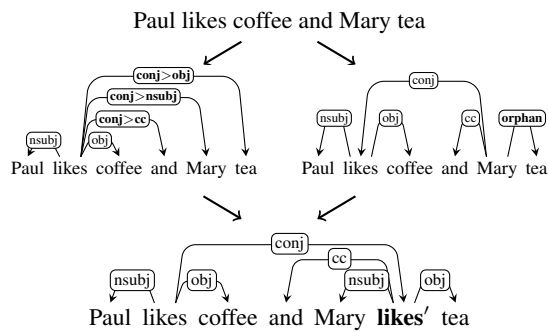

While gapping constructions receive a lot of attention in the theoretical syntax literature
(e.g., \citealt{Ross1970,Jackendoff1971,Steedman1990,Coppock2001,Osborne2006,Johnson2014,Toosarvandani2016,Kubota2016}),
they have been almost entirely neglected by the NLP community so far. The Penn Treebank explicitly annotates gapping constructions,
by co-indexing arguments in the clause with a predicate and the clause with the gap, but these co-indices 
are not included in the standard parsing metrics and almost all parsers ignore them.\footnote{
To the best of our knowledge, the parser by Kummerfeld and Klein (\citeyear{Kummerfeld2017}) is the only 
parser that tries to output the co-indexing of constituents in clauses with gapping but they 
lack an explicit evaluation of their co-indexing prediction accuracy.} 
Despite the sophisticated analysis of gapping within CCG \cite{Steedman1990},
sentences with gapping were deemed too difficult to represent within the CCGBank \cite{Hockenmaier2007}. 
Similarly the treebanks for the Semantic Dependencies Shared Task 
\cite{Oepen2015} exclude all sentences from the Wall Street Journal that contain gapping. 
Finally, while the tectogrammatical layer of the Prague Dependency Treebank \cite{PDT2013} as well as the 
enhanced Universal Dependencies (UD) representation \cite{Nivre2016} provide an analysis with reconstructed nodes
for gapping constructions, there exist no methods to automatically parse to these representations.

Here, we provide the first careful analysis of parsing of gapping constructions,
and we present two methods for reconstructing elided predicates in sentences with
gapping within the UD framework. As illustrated in Figure~\ref{fig:overview},
we first parse to a dependency tree and then reconstruct the elided material.
The methods differ in how much information is encoded in the dependency tree. The first method
adapts an existing procedure for parsing sentences with elided function words \cite{Seeker2012}, 
which uses composite labels that can be deterministically turned into dependency graphs in most cases. The
second method is a novel procedure that relies on the parser 
only to identify a gap, and then employs an unsupervised method to reconstruct 
the elided predicates and reattach the arguments to the reconstructed predicate. We find that
both methods can reconstruct elided predicates with very high accuracy from gold standard
dependency trees. When applied to the output of a parser, which often fails to identify gapping, 
our methods achieve a sentence-level accuracy of 32\% and 34\%, significantly outperforming
the recently proposed constituent parser by \citet{Kummerfeld2017}.

\section{Background}

\subsection{Gapping constructions}
\label{sec:gapping-constructions}
Gapping constructions in English come in many forms that can be broadly classified as follows.
\ex. \label{ex:single-verb} 
       {\it Single predicate gaps}: \\
       John {\bf bought} books, and Mary \underline{{ }{ }{ }{ }{ }} flowers. 
       
\ex. {\it Contiguous predicate-argument gap \\ (including ACCs)}: \\
       Eve {\bf gave flowers} to Al and Sue  \underline{{ }{ }{ }{ }{ }}  to Paul. \\
       Eve {\bf gave} a CD to Al and  \underline{{ }{ }{ }{ }{ }} roses to Sue. 

\ex. {\it Non-contiguous predicate-argument gap}: \\
       Arizona {\bf elected} Goldwater {\bf Senator}, and Pennsylvania  \underline{{ }{ }{ }{ }{ }} Schwelker  \underline{{ }{ }{ }{ }{ }}. \\
       \null \hfill \cite{Jackendoff1971}
       
\ex.  \label{ex:non-finite} 
       {\it Verb cluster gap}: \\
       I {\bf want to try to begin to write a novel} and  \\
       \null \hfill ... Mary   \underline{{ }{ }{ }{ }{ }} a play. \\
       \null \hfill ... Mary   \underline{{ }{ }{ }{ }{ }} to write a play. \\
       \null \hfill ... Mary   \underline{{ }{ }{ }{ }{ }} to begin to write a play. \\
       \null \hfill ... Mary   \underline{{ }{ }{ }{ }{ }} to try to begin to write a play. \\
       \null \hfill \cite{Ross1970}

The defining characteristic of gapping constructions is that there is a clause 
that lacks a predicate (the {\it gap}) but still contains two or more arguments or modifiers 
of the elided predicate (the {\it remnants} or {\it orphans}). In most cases, the remnants have
a corresponding argument or modifier (the {\it correspondent}) in the clause with the overt
predicate.

These types of gapping also make up the 
majority of attested constructions in other languages.
However, Wyngaerd (\citeyear{Wyngaerd2007}) notes that Dutch
permits gaps in relative clauses, and Farudi (\citeyear{Farudi2013}) notes that Farsi permits 
gaps in finite embedded clauses even if the overt predicate is not embedded.\footnote{See Johnson (\citeyear{Johnson2014}) or Schuster et al. (\citeyear{Schuster2017})
for a more comprehensive overview of cross-linguistically attested gapping constructions. }

\subsection{Target representation}

We work within the UD framework, which aims 
to provide cross-linguistically consistent dependency annotations that are useful for NLP tasks. 
UD defines two types of representation: the {\it basic} UD representation which
is a strict surface syntax dependency tree and the {\it enhanced} UD
 representation \citep{Schuster2016} which may be a graph instead of a tree and may contain additional nodes. 
The analysis of gapping in the enhanced representation makes use of copy nodes 
for elided predicates and additional edges for elided arguments, which we both try
to automatically reconstruct in this paper. In the simple case in which only one predicate was elided,
there is exactly one copy node for the elided predicate, which leads to a structure that is 
identical to the structure of the same sentence without a gap.\footnote{To 
enhance the readability of our examples, we place the copy node in 
the sentence where the elided predicate would have been pronounced. 
However, as linear order typically does not matter for extracting information 
with dependency patterns, our procedures only try to recover the structure of 
canonical sentences but not their linear order.} 
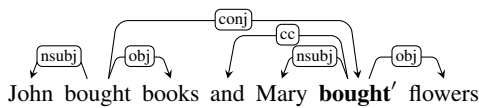
\begin{center}
\footnotesize
\begin{dependency}
\begin{deptext}
John \& bought \& books \& and \& Mary \& {\bf bought$'$} \& flowers \\
\end{deptext}
\depedge[edge unit distance=2ex]{2}{1}{nsubj}
\depedge[edge unit distance=2ex]{2}{3}{obj}
\depedge[edge unit distance=2ex]{6}{4}{cc}
\depedge[edge unit distance=2ex]{6}{5}{nsubj}
\depedge[edge unit distance=1.35ex]{2}{6}{conj}
\depedge[edge unit distance=2ex]{6}{7}{obj}
\end{dependency}
\end{center}

If a clause contains a more complex gap, the enhanced representation contains copies for
all content words that are required to attach the remnants.

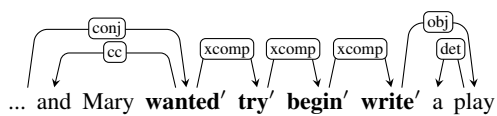
\begin{center}
\footnotesize
\begin{dependency}
\begin{deptext}
... \& and \& Mary \& {\bf wanted$'$} \& {\bf try$'$}  \& {\bf begin$'$} \&  {\bf write$'$} \& a \& play  \\
\end{deptext}
\depedge[edge unit distance=1.6ex]{4}{2}{cc}
\depedge[edge unit distance=1.8ex]{1}{4}{conj}
\depedge{4}{5}{xcomp}
\depedge{5}{6}{xcomp}
\depedge{6}{7}{xcomp}
\depedge{9}{8}{det}
\depedge[edge unit distance=3ex]{7}{9}{obj}
\end{dependency}
\end{center}

The motivation behind this analysis is that the semantically empty markers {\it to} 
are not needed for interpreting the sentence and minimizing the number of
copy nodes leads to less complex graphs.

Finally, if a core argument was elided along with the predicate, we introduce additional
dependencies between the copy nodes and the shared arguments, as
for example, the open clausal complement ({\tt xcomp}) dependency between the copy node
and {\it Senator} in the following example.

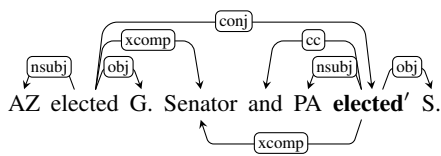
\begin{center}
\footnotesize
\begin{dependency}
\begin{deptext}
AZ \& elected \& G. \& Senator \& and \& PA \& {\bf elected$'$} \& S. \\
\end{deptext}
\depedge[edge unit distance=2ex]{2}{1}{nsubj}
\depedge[edge unit distance=2ex]{2}{3}{obj}
\depedge[edge unit distance=2.25ex]{2}{4}{xcomp}
\depedge[edge unit distance=2.25ex]{7}{5}{cc}
\depedge[edge unit distance=2ex]{7}{6}{nsubj}
\depedge[edge unit distance=1.25ex]{2}{7}{conj}
\depedge[edge unit distance=2ex]{7}{8}{obj}
\depedge[edge below,edge unit distance=0.75ex]{7}{4}{xcomp}
\end{dependency}
\end{center}

The rationale for not copying all arguments is again to keep the graph simple, while still
encoding all relations between content words. Arguments can be arbitrarily complex 
and it seems misguided to copy entire subtrees of arguments which, e.g.,  could contain 
multiple adverbial clauses. Note that linking to existing nodes would not work in the case 
of verb clusters because they do not satisfy the subtree constraint.

\section{Methods}

\subsection{Composite relations}

Our first method adapts one of the procedures by \citet{Seeker2012}, which represents
gaps in dependency trees by attaching dependents of an elided predicate with composite relations.  These relations 
represent the dependency path that would  have existed if nothing had been elided. For example, in the following sentence, 
the verb {\it bought}, which would have been attached to the head of the first conjunct with a {\tt conj} relation, was elided from 
the second conjunct and hence all nodes that would have depended on the elided verb, are attached to the first conjunct using a composite
relation consisting of {\tt conj} and the type of argument.
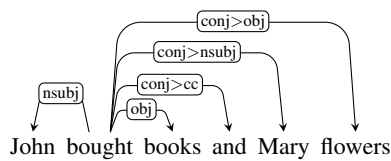
\begin{center}
\footnotesize
\begin{dependency}
\begin{deptext}
John \& bought \& books \& and \& Mary \& flowers \\
\end{deptext}
\depedge{2}{1}{nsubj}
\depedge[edge unit distance=2ex]{2}{3}{obj}
\depedge[edge unit distance=2.15ex]{2}{4}{conj$>$cc}
\depedge[edge unit distance=2.45ex]{2}{5}{conj$>$nsubj}
\depedge[edge unit distance=2.55ex]{2}{6}{conj$>$obj}
\end{dependency}
\end{center}

The major advantage of this approach is that the dependency tree contains information about the types of arguments
and so it should be straightforward to turn dependency trees of this form into enhanced UD graphs. For most dependency trees, one can
obtain the enhanced UD graph by splitting the composite relations into its atomic parts and inserting
copy nodes at the splitting points.\footnote{Note that this 
representation does not indicate conjunct boundaries, and for sentences with multiple gapped conjuncts,
 it is thus unclear how many copy nodes are required.}

At the same time, this approach comes with the drawback of drastically increasing the label space.
For sentences with more complex gaps as in \ref{ex:non-finite}, one has to use composite relations that consist of more than two
atomic relations and theoretically, the number of composite relations is unbounded:
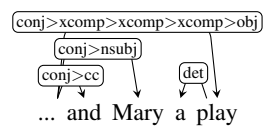
\begin{center}
\footnotesize
\begin{dependency}
\begin{deptext}
... \& and \& Mary \& a \& play  \\
\end{deptext}
\depedge[edge unit distance=2.25ex]{5}{4}{det}
\depedge[edge unit distance=1.75ex]{1}{5}{conj$>$xcomp$>$xcomp$>$xcomp$>$obj}
\depedge[edge unit distance=2.25ex]{1}{3}{conj$>$nsubj}
\depedge[edge unit distance=2ex]{1}{2}{conj$>$cc}
\end{dependency}
\end{center}
\subsection{Orphan procedure}

Our second method also uses a two-step approach to resolve gaps, but compared to the previous method,
it puts less work on the parser. We first parse sentences to the basic UD v2 representation, which analyzes
gapping constructions as follows. One remnant is promoted to be the head of the clause
and all other remnants are attached to the promoted phrase. For example, in this
sentence, the subject of the second clause, {\it Mary}, is the head of the
clause and the other remnant, {\it flowers}, is attached to {\it Mary} with the special {\tt orphan}
relation:

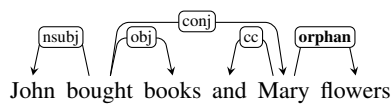
\begin{center}
\footnotesize
\begin{dependency}
\begin{deptext}
John \& bought \& books \& and \& Mary \& flowers \\
\end{deptext}
\depedge{2}{1}{nsubj}
\depedge{2}{3}{obj}
\depedge{5}{4}{cc}
\depedge[edge unit distance=1.65ex]{2}{5}{conj}
\depedge{5}{6}{\bf orphan}
\end{dependency}
\end{center}

This analysis can also be used for more complex gaps, as in the example
with a gap that consists of a chain of non-finite embedded verbs in \ref{ex:non-finite}.

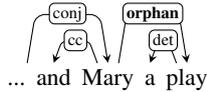
\begin{center}
\footnotesize
\begin{dependency}
\begin{deptext}
... \& and \& Mary \& a \& play  \\
\end{deptext}
\depedge[edge unit distance=2ex]{3}{2}{cc}
\depedge[edge unit distance=2.25ex]{1}{3}{conj}
\depedge[edge unit distance=2ex]{5}{4}{det}
\depedge[edge unit distance=2.25ex]{3}{5}{\bf orphan}
\end{dependency}
\end{center}

When parsing to this representation, the parser only has to identify that there is a gap
 but does not have to recover the elided material or determine the type of remnants.
As a second step, we use an unsupervised procedure to determine which nodes
to copy and how and where to attach the remnants. In developing this 
procedure, we made use of the fact that in the vast majority of cases, 
all arguments and modifiers that are expressed in gapped conjunct are 
also expressed in the full conjunct. The problem of determining which nodes to copy and 
which relations to use can thus be reduced to the problem of aligning arguments 
in the gapped conjunct to arguments in the full conjunct. We apply the following procedure 
to all sentences that contain at least one {\tt orphan} relation.
\begin{enumerate}
\itemsep0em 
\item Create a list $F$ of arguments of the head of the full conjunct by considering all core argument dependents of the conjunct's head as well as clausal and nominal non-core dependents, and adverbial modifiers.
\item Create a list $G$ of arguments in the gapped conjunct that contains the head of the gapped conjunct and all its {\tt orphan} dependents.
\item Find the highest-scoring monotonic alignment of arguments in $G$ to arguments in $F$.
\item Copy the head of the full conjunct and attach the copy node $c$ to the head of the full conjunct with the original relation of the head of the gapped conjunct (usually {\tt conj}).
\item For each argument $g \in G$ that has been aligned to $f \in F$, attach $g$ to $c$ with the same relation as the parent relation of $f$, e.g., if $f$ is attached to the head of the full conjunct with an {\tt nsubj} relation, also attach $g$ to $c$ with an {\tt nsubj} relation. Attach arguments $g' \in G$ that were not aligned to any token in $F$ to $c$ using the general {\tt dep} relation.
\item For each copy node $c$, add dependencies to all core arguments of the original node which do not have a corresponding remnant in the gapped clause. For example, if the full conjunct contains a subject, an object, and an oblique modifier but the clause with the gap, only a subject and an oblique modifier, add an object dependency between the copy node and the object in the full conjunct.
\end{enumerate}

\begin{table*}
\footnotesize
\begin{tabularx}{\textwidth}{l c c c c  c c c c c c c }
& \multicolumn{3}{ c }{ \textsc{ewt}} & \ \  \ & \multicolumn{3}{ c }{   \textsc{gapping}} &  \  \ \ & \multicolumn{3}{ c }{ \textsc{ewt + gapping} (\textsc{combined})} \\
& {\bf Train} & {\bf Dev} & {\bf Test} & & {\bf Train} & {\bf Dev} &  {\bf Test}& & {\bf Train} & {\bf Dev} & {\bf Test}\\ \midrule
sentences & 12,543 & 2,002 &2,077 && 164  & 79 & 79 &&  12,707 & 2,081 & 2,156\\ \midrule
tokens & 204,585 & 25,148 & 25,096  && 4,698 & 2,383 & 2,175 && 209,283 & 27,531 & 27,271\\ \midrule
{sentences with} & 21 & 1 & 1& & 164 & 79 &79 && 185 & 80 & 80\\
 gapping & 0.15\% & 0.05\% & 0.05\% &&  100\% & 100\% & 100\% && 1.46\% & 3.84\%  & 3.71\% \\ \midrule
{copy nodes } & 22 & 2 & 1 && 201 & 96 & 102 && 223 & 98 & 103\\ \midrule
{unique composite} \ \  &16 & 6 & 2 && 41 & 29 & 31 && 46 & 29 & 31  \\
relations  & & & & & & & & & &  & \\ 
\end{tabularx}
\caption{Treebank statistics. The \textit{copy nodes} row lists the number of copy nodes and the \textit{unique composite relations} row lists the number of unique composite relations in the treebanks annotated according to the \textsc{composite} analysis. The percentages are relative to the total number of sentences.}\label{tbl:data-stats}
\end{table*}

\begin{table}
\footnotesize
\begin{tabularx}{\columnwidth}{l c }
\textbf{Gap type} & \textbf{Frequency} \\ \midrule
Single predicate & 172\\
Contiguous predicate-argument & 140\\
Non-contiguous predicate-argument & 9 \\
Verb cluster & 24 \\
\end{tabularx}
\caption{Distribution of gap types in our corpus. The classification is according to the four types of gaps that we discussed in Section~\ref{sec:gapping-constructions}.}\label{tbl:gapping-types}
\end{table}

A crucial step is the third step, determining the highest-scoring alignment. This can be done straightforwardly with the sequence alignment algorithm by \citet{Needleman1970} if one defines a similarity function $sim(g, f)$ that returns a similarity score between the arguments $g$ and $f$. We defined $sim$ based on the intuitions that often, parallel arguments are of the same syntactic category, that they are introduced by the same function words (e.g., the same preposition), and that they are closely related in meaning. The first intuition can be captured by penalizing mismatching POS tags, and the other two by computing the distance between argument embeddings. We compute these embeddings by averaging over the 100-dim.\ pretrained GloVe \cite{Pennington2014} embeddings for each token in the argument. Given the POS tags $t_{g}$ and $t_{f}$ and the argument embeddings $v_{g}$ and $v_{f}$, $sim$ is defined as follows.\footnote{As suggested by one of the reviewers, we also ran a post-hoc experiment with a simpler similarity score function without the embedding distance term, which only takes into account whether the POS tags match. We found that quantitatively, the embeddings do not lead to significant better scores on the test set according to our metrics but qualitatively, they lead to better results for the examples with verb cluster gaps.}
\begin{align*}
sim(g, f) = -
 \lVert v_{g}& -v_{f} \rVert_2  + \mathbbm{1}\left[t_{g} = t_{f} \right] \\ 
 & \times pos\_mismatch\_penalty
  \end{align*}

We set $pos\_mismatch\_penalty$, a parameter that penalizes mismatching POS tags, to $-2$.\footnote{We optimized this parameter on the training set by trying integer values from $-$1 to $-$15.}


This procedure can be used for almost all sentences with gapping constructions. However, if parts of an argument were elided along with the main predicate, it can become necessary to copy multiple nodes. We therefore consider the alignment not only between complete arguments in the full clause and the gapped clause but also between partial arguments in the full clause and the complete arguments in the gapped clause. For example, for the sentence ``{\it Mary wants to write a play and Sue a book}'' the complete arguments of the full clause are \{{\it Mary}, {\it to write a play}\} and   the arguments of the gapped clause are \{{\it Sue}, {\it a book}\}. In this case, we also consider the partial arguments  \{{\it Mary}, {\it a play}\} and if the arguments of the gapped conjunct align better to the partial arguments, we use this alignment. However, now that the token {\it write} is part of the dependency path between {\it want} and {\it play}, we also have to make a copy of {\it write} to reconstruct the UD graph of the gapped clause.

\section{Experiments}

Both methods rely on a dependency parser followed by 
a post-processing step. We evaluated the individual steps
and the end-to-end performance.


\subsection{Data}

We used the UD English Web Treebank v2.1 \cite[henceforth \textsc{ewt};][]{Silveira2014,Nivre2017a} for training and evaluating parsers. As the treebank is relatively small and therefore only contains very few sentences with gapping, we also extracted gapping constructions from the WSJ and Brown portions of the PTB \citep{Marcus1993} and the GENIA corpus \citep{Ohta2002}. Further, we copied sentences from the Wikipedia page on gapping\footnote{\url{https://en.wikipedia.org/wiki/Gapping}, accessed on Aug 24, 2017.} and from published papers on gapping. The sentences in the \textsc{ewt} already contain annotations with the {\tt orphan} relation and copy nodes for the enhanced representation, and we manually added both of these annotations for the remaining examples. The composite relations can be automatically obtained from the enhanced representation by removing the copy nodes and concatenating the dependency labels, which we did to build the training and test corpus for the composite relation procedure. 
Table~\ref{tbl:data-stats} shows properties of the data splits of the original treebank, the additional sentences with gapping, and their combination; Table~\ref{tbl:gapping-types} shows the number of sentences in our corpus for each of the gap types.

\subsection{Parsing experiments}

\paragraph{Parser} We used the parser by Dozat and Manning \shortcite{Dozat2017} 
for parsing to the two different intermediate dependency representations.
This parser is a graph-based parser \cite{McDonald2005} that uses a biLSTM to compute token representations
and then uses a multi-layer perceptron with biaffine attention to compute arc and label scores. 

\paragraph{Setup} We trained the parser on the {\sc combined} training corpus with gold tokenization, and predicted fine-grained and universal part-of-speech tags, for which we used the tagger by \citet{Dozat2017b}. We trained the tagger on the \textsc{combined} training corpus. As pre-trained embeddings, we used the word2vec \cite{Mikolov2013} embeddings that were provided for the CoNLL 2017 Shared Task \cite{Zeman2017}, and we used the same hyperparameters as \citet{Dozat2017b}.

\paragraph{Evaluation} We evaluated the parseability of the two dependency representations using labeled and unlabeled attachment scores (LAS and UAS). Further, to specifically evaluate how well parsers are able to parse gapping constructions according to the two annotation schemes, we also computed the LAS and UAS just for the head tokens of remnants (LAS$_{g}$ and UAS$_{g}$). For all our metrics, we excluded punctuation tokens. To determine statistical significance of pairwise comparisons, we performed two-tailed approximate randomization tests \cite{Noreen1989,Yeh2000} with an adapted version of the {\tt sigf} package \citep{Pado2006}.

\begin{table}
\footnotesize
\begin{tabularx}{\columnwidth}{l  l   c c c l }
		&			&	\multicolumn{2}{c}{\textsc{ewt}} 	&	\multicolumn{2}{c}{\textsc{gapping} \ \ \ \ \ } \\
	&				&	UAS&	LAS&	UAS&	LAS \ \ \ \ \ \ \ \ \ 	 	 \\\midrule
\multirow{2}{*}{\bf Dev} &	\textsc{orphan}&	\bf{90.57}&	{87.32}&	{\bf 89.34} &		{\bf 85.69**}	 \\
&	\textsc{composite} &		90.46&	{\bf87.37}&	88.86&	84.21		\\\midrule
\end{tabularx}

\begin{tabularx}{\columnwidth}{ l l   c c c l }
\multirow{2}{*}{\bf Test}  &	\textsc{orphan}&	  {90.42}&	{87.06}&	{\bf 87.44}&	{\bf 83.97**}\\
& 	\textsc{composite}   &	        {\bf 90.54} &	{\bf 87.33} &	{86.51}&	{81.69\ \ }			\\
\end{tabularx}

\caption{\label{tbl:parsing-results} Labeled (LAS) and unlabeled attachment score (UAS) of parsers trained and evaluated on the UD representation ({\sc orphan}) and the composite relations representation ({\sc composite}) on the development and test sets of the {\sc ewt} and the {\sc gapping} treebank.  **~indicates that results differ significantly  at $p<0.01$.}
\end{table}

\begin{table}
\footnotesize

\begin{tabularx}{\columnwidth}{ l   l l l l }
					&	\multicolumn{2}{c}{\textbf{Development}} 	&	\multicolumn{2}{c}{\textbf{Test} } 		 \\
					&	UAS$_g$&	LAS$_g$&	UAS$_g$&	LAS$_g$	 	 \\\midrule
	\textsc{orphan}&	\bf{72.36}&	{\bf64.73***}&	{\bf 72.56}* &		{\bf 65.79***}	 \\
	\textsc{composite} &		68.36&	{49.45}&	62.41&	46.24		\\
\end{tabularx}
\caption{\label{tbl:parsing-results-g} Labeled (LAS$_g$) and unlabeled attachment score (UAS$_g$) of head tokens of remnants for parsers trained and evaluated on the UD representation ({\sc orphan}) and the composite relations representation ({\sc composite}) on the development and test sets of the {\sc combined}  treebank. Results that differ significantly are marked with  *~($p<0.05$) or ***~($p<0.001$).}
\end{table}

\paragraph{Results} 

Table~\ref{tbl:parsing-results} shows the overall parsing results on the development and test sets of the two treebanks. 
There was no significant difference between the parser that was trained on the UD representation ({\sc orphan})
and the parser trained on the composite representation ({\sc composite}) when tested on the {\sc EWT} data sets, which is 
not surprising considering that there is just one sentence with gapping each in the development and the test split. When evaluated
on the {\sc gapping} datasets, the {\sc orphan} parser performs significantly better ($p<0.01$) in terms of labeled attachment score,
which suggests that the parser trained on the {\sc composite} representation is indeed struggling with the greatly increased label space.
This is further confirmed by the attachment scores of the head tokens of remnants (Table~\ref{tbl:parsing-results-g}). 
The labeled attachment score of remnants is significantly higher for the {\sc orphan} parser than for the {\sc composite} parser.
Further, the unlabeled attachment score on the test set is also higher for the {\sc orphan} parser, which suggests that the {\sc composite} parser is
sometimes struggling with finding the right attachment for the multiple long-distance composite dependencies.

\begin{table*}
\footnotesize
\begin{tabularx}{\textwidth}{ l  l   l l l l l  l l l l l }
& & \multicolumn{5}{c }{\bf{Development} \ \ \ \ \ \ \ \ \  } &  \multicolumn{5}{c}{\bf{Test}  \ \ \ \ \  } \\ \midrule
&					&	UP&	UR&	LP & LR & SAcc. 	& 	UP &	UR&	LP & LR & SAcc.  	 \\\midrule
\multirow{2}{*}{oracle} &	\textsc{composite}&	{91.32}&	{88.20}&	{\bf 91.32} &		{\bf 88.20}	 & {\bf91.14**}&	{90.71}&	{86.81}&	{\bf 90.71} & {86.81} &  {\bf 86.08*} \\
&	\textsc{orphan}  &		\bf{94.08}&	{\bf93.79}&	87.54&	87.27&	         {72.15} &	{\bf 92.02} &	{\bf 92.02}&	{87.12} 	 & {\bf 87.12}  & {72.15}		\\ \midrule \midrule
\multirow{2}{*}{end-to-end} &	\textsc{composite}&	{70.48}&	{\bf 49.69*}&	{\bf 65.64} &		{\bf 46.27*}	 & {\bf 31.65}&	{67.39}&	{\bf 47.55}&	{61.74} & {\bf 43.56} &  {31.65} \\
 &	\textsc{orphan}  &		\bf{71.73}&	{42.55}&	65.45&	38.82&	         {30.38} &	{\bf 78.92**} &	{44.78}&	{\bf 68.11} 	 & {38.65}  & {\bf 34.18}		\\ \midrule \midrule
& 	K\&K 2017 & - & - & - & - & \hphantom{0}0.00 & - & - & - & - & {\hphantom{0}0.00} 

\end{tabularx}
\caption{\label{tbl:enhancement-results} Labeled and unlabeled precision and recall as well as sentence-level accuracy of the two gapping reconstructions methods and the K\&K parser on the development and test set of the {\sc combined} treebank.  Results that differ significantly from the other result within the same section are marked with  *~($p<0.05$) or **~($p<0.01$).}
\end{table*}

\subsection{Recovery experiments}

Our second set of experiments concerns the recovery of the elided material and the reattachment of the orphans.
We conducted two experiments: an oracle experiment that used gold standard dependency trees
and an end-to-end experiment that used the output of the parser as input. For all experiments, we used the 
{\sc combined} treebank.

\paragraph{Evaluation} Here, we evaluated dependency graphs and therefore 
used the labeled and unlabeled precision and recall metrics. However, as our two procedures
are only changing the attachment of orphans, we only computed these metrics for copy nodes and their
dependents. Further, we excluded punctuation and coordinating conjunctions
as their attachment is usually trivial and including them would inflate scores. Lastly, we 
computed the sentence-level accuracy for all sentences with gapping. For this metric, we considered a sentence
to be correct if all copy nodes and their dependents of a sentence were attached to the correct head with the correct label.

\paragraph{Oracle results}

The top part of Table~\ref{tbl:enhancement-results} shows the results for the oracle experiment. 
Both methods are able to reconstruct the elided material and the canonical clause structure from gold dependency trees 
with high accuracy. This was expected for the  {\sc composite} procedure, which can make use of the composite relations in the 
dependency trees, but less so for the {\sc orphan} procedure which has to recover the structure and the types of relations.
The two methods work equally well in terms of all metrics except for the sentence-level accuracy, which is significantly higher
for the {\sc composite} procedure. This difference is caused by a difference in the types of mistakes.
All errors of the {\sc composite} procedure are of a structural nature and stem from copying the wrong number of nodes while
the dependency labels are always correct because they are part of the dependency tree.
The majority of errors of the {\sc orphan} procedure stem from incorrect dependency labels, and these mistakes are
scattered across more examples, which leads to the lower sentence-level accuracy.

\paragraph{End-to-end results}
The middle part of Table~\ref{tbl:enhancement-results} shows the results for the end-to-end experiment. 
The performance of both methods is considerably lower than in the oracle experiment, which is primarily driven
by the much lower recall. Both methods assume that the parser detects the existence of a gap and if the parser
fails to do so, neither method attempts to reconstruct the elided material. In general, precision tends to be a bit
higher for the {\sc orphan} procedure whereas recall tends to be a bit higher for the {\sc composite} method but overall
and in terms of sentence-level accuracy both methods seem to perform equally well.

\paragraph{Error analysis}
For both methods, the primary issue is low recall, which is a result of parsing errors.
When the parser correctly predicts the {\tt orphan} relation, the main
sources of error for the {\sc orphan} procedure are missing correspondents
for remnants (e.g., {\it [for good]} has no correspondent in 
 {\it They had left the company, many for good}) or that the types of argument 
 of the remnant and its correspondent differ (e.g., in  {\it She was convicted 
 of selling unregistered securities in Florida and of unlawful phone calls in Ohio},
 {\it  [of selling unregistered securities] }  is an 
 adverbial clause whereas {\it [of unlawful phone calls]} is an oblique modifier).
 
Apart from the cases where the {\sc composite} procedure leads to an incorrect structure,
the remaining errors are all caused by the parser predicting the wrong composite relation.
 
\subsection{Comparison to Kummerfeld and Klein}

\citet[henceforth K\&K;][]{Kummerfeld2017} recently proposed a one-endpoint-crossing graph parser that
is able to directly parse to PTB-style trees with traces. They also briefly discuss gapping
constructions and their parser tries to output the co-indexing that is used for gapping
constructions in the PTB. The EWT and all the sentences that we took
from the WSJ, Brown, and GENIA treebanks already come with constituency
tree annotations, and we manually annotated the remaining sentences according to the 
PTB guidelines \cite{Bies1995}. This allowed us to train the  K\&K parser with exactly
the same set of sentences that we used in our previous experiments. 
As this parser outputs constituency trees, we could not compute dependency
graph metrics for this method. For the sentence-level accuracy, we considered an example to be correct if
a) each argument in the gapped conjunct was the child of a single constituent node, which in return was the sibling of the full clause/verb phrase,
and b) the co-indexing of each argument in the gapped conjunct was correct. For example, the following bracketing
would be considered correct despite the incorrect internal structure of the first conjunct:

{ \footnotesize
\noindent $ [_\textnormal{S} [_\textnormal{S} [_{\textnormal{NP-1}} \mbox{ Al }] \mbox{ likes } [_{\textnormal{NP-2}} \mbox{ coffee }] ] \mbox{ and } [_{\textnormal{S}} [_{\textnormal{NP=1}} \mbox{ Sue }] [_{\textnormal{NP=2}} \mbox{ tea }] ]]$}

The last row of Table~\ref{tbl:enhancement-results} shows the results of the K\&K parser. The parser failed to output the correct constituency structure or co-indexing for every single example in the development and test sets.
The parser struggled in particular with outputting the correct co-indices:
For 32.5\% of the test sentences with gapping, the bracketing of the gapped clause
was correct but one or more of the co-indices were missing from the output.

Overall these results suggest that our depend-ency-based approach
is much more reliable at identifying gapping constructions than the parser by K\&K, which, in their defense, was 
optimized to output traces for other phenomena. Our method is also faster
and took only seconds to parse the test set, while the K\&K parser took several hours.

\section{Resolving gaps in other languages}

One of the appeals of the {\sc orphan} procedure is that it can be easily applied to other languages even
if there exist no annotated enhanced dependency graphs.\footnote{There 
is no theoretical reason that would prevent one from using the \textsc{composite} procedure for other languages but
given that UD treebanks are annotated with \texttt{orphan} relations, using the  the \textsc{composite} procedure would require additional manual annotations in practice. }
On the one hand, this is because our method does not make use of lexical information,
and on the other hand, this is because we developed our method on top of the UD
annotation scheme, which has already been applied to many languages and for which many treebanks exist.

\begin{figure*}
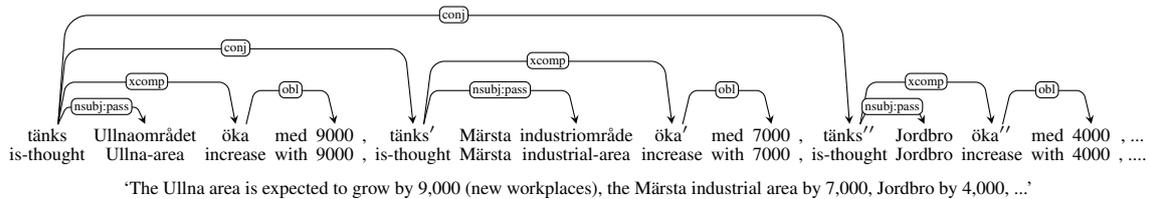

\begin{center}
\scriptsize
\begin{dependency}
\begin{deptext}
t\"{a}nks \& Ullnaomr{\aa}det \& \"oka \& med \& 9000 \& , \& t\"anks$'$ \& M\"arsta \&  industriomr{\aa}de \& \"oka$'$ \&   med \& 7000 \& , \& t\"anks$''$ \& Jordbro \&  \"oka$''$ \& med \& 4000 \& ,  ... \\
is-thought \& Ullna-area \& increase \& with \& 9000 \& , \& is-thought \& M\"arsta \& industrial-area \& increase \& with \& 7000 \& , \& is-thought \& Jordbro \& increase \& with \& 4000 \&, .... \\
\end{deptext}

\depedge[edge unit distance=2ex]{1}{2}{nsubj:pass}
\depedge[edge unit distance=2.5ex]{1}{3}{xcomp}
\depedge[edge unit distance=2ex]{3}{5}{obl}
\depedge[edge unit distance=1.5ex]{1}{7}{conj}
\depedge[edge unit distance=2ex]{7}{9}{nsubj:pass}
\depedge[edge unit distance=2.5ex]{7}{10}{xcomp}
\depedge[edge unit distance=2ex]{10}{12}{obl}
\depedge[edge unit distance=1ex]{1}{14}{conj}
\depedge[edge unit distance=2.5ex]{14}{16}{xcomp}
\depedge[edge unit distance=2ex]{14}{15}{nsubj:pass}
\depedge[edge unit distance=2ex]{16}{18}{obl}

\end{dependency}

`The Ullna area is expected to grow by 9,000 (new workplaces), the M\"arsta industrial area by 7,000, Jordbro by 4,000, ...'
\end{center}
\caption{\label{fig:swedish-example} Dependency graph for part of the sentence \texttt{sv-ud-train-1102} as output by the \textsc{orphan} procedure. The system correctly predicts the copy nodes for the matrix and the embedded verb, and correctly attaches the arguments to the copy nodes.}
\end{figure*}

Currently, all treebanks but the English one lack copy nodes for gapping constructions and many of them incorrectly use the 
{\tt orphan} relation \cite{Droganova2017} and therefore we could
not evaluate our method on a large variety of languages. In order to demonstrate that our method can be applied
to other languages, we therefore did a case study on the Swedish UD treebank. The Swedish UD
treebank is an automatic conversion from a section of the Talbanken \cite{Einarsson1976} with extensive manual corrections.
While the treebank is overall of high quality, we noticed conversion errors that led to incorrect uses of the {\tt orphan}
relation in 11 of the 29 sentences with {\tt orphan} relations, which we excluded from our evaluation. 
We applied our gapping resolution procedure without any modifications to the remaining 18 sentences.
We used the Swedish word2vec embeddings that were prepared for the CoNLL 2017
Shared Task. Our method correctly predicts the insertion of 29 copy nodes and is able to predict the correct
structure of the enhanced representation in all cases, including complex ones with elided verb clusters such as the example in Figure~\ref{fig:swedish-example}. 
It also predicts the correct dependency label for 108/110 relations,
leading to a labeled precision and labeled recall of 98.18\%, which are both higher than the English numbers despite the
fact that we optimized our procedure for English. 
The main reason for the higher performance seems to be that many of the Swedish examples come from
informational texts from public organizations, which are more likely to be written to be clear and unambiguous.
Further, the Swedish data does not contain challenging examples from the linguistic literature. 

As Swedish is a Germanic language like English and thus shares many structural properties, 
we cannot conclude that our method is applicable to any language based 
on just this experiment. However, given 
that our method does not rely on language-specific structural patterns, we expect it
to work well for a wide range of languages.

\section{Related work}

Gapping constructions have been little studied in NLP, but several approaches (e.g., \citealt{Dukes2011,Simko2017}) parse to dependency trees with empty nodes. \citet{Seeker2012} compared three ways of parsing with empty heads: adding a transition that inserts empty nodes, using composite relation labels for nodes that depend on an elided node, and pre-inserting empties before parsing. These papers all focus on recovering nodes for elided function words such as auxiliaries; none of them attempt to recover and resolve the content word elisions of gapping. \citet{Ficler2016} modified PTB annotations of argument-cluster coordinations (ACCs), i.e., gapping constructions with two post-verbal orphan phrases, which make up a subset of the gapping constructions in the PTB\@. While the modified annotation style leads to higher parsing accuracy of ACCs, it is specific to ACCs and does not generalize to other gapping constructions. Moreover, they did not reconstruct gapped ACC clauses.  Traditional grammar-based chart parsers \cite{Kay1980,Klein2001} did handle empty nodes and so could in principle provide a parse of gapping sentences though additional mechanisms would be needed for reconstruction. In practice, though, dealing with gapping in a grammar-based framework is not straightforward and can lead to a combinatorial explosion that slows down parsing in general, as has been noted for the English Resource Grammar (Flickinger, 2017, p.c.) and for an HPSG implementation for Norwegian \cite{Haugereid2017}. The grammar-based parser built with augmented transition networks \cite{Woods1970} provided an extension in the form of the SYSCONJ operation \cite{Woods1973} to parse some gapping constructions, but also this approach lacked explicit reconstruction mechanisms and provided only limited coverage.

There also exists a long line of work on post-processing surface-syntax constituency trees to recover traces in the PTB 
 \cite{Johnson2002,Levy2004, Campbell2004,Gabbard2006}, pre-processing sentences such that they contain tokens for traces
  before parsing \cite{Dienes2003}, or directly parsing sentences to either PTB-style trees with empty elements or 
  pre-processed trees that can  be deterministically converted to PTB-style trees
 \cite{Collins1997,Dienes2003b,Schmid2006,Cai2011,Hayashi2016,Kato2016,Kummerfeld2017}. 
However, all of these works are primarily concerned with recovering traces for phenomena such as Wh-movement or control and raising constructions and, 
with the exception of \citet{Kummerfeld2017}, none of these works attempt to output the co-indexing that is 
used for analyzing gapping constructions. And again, none of these works try to reconstruct elided material.

Lastly,  several methods have been proposed for resolving other forms of ellipsis, including VP ellipsis 
\cite{Hardt1997,Nielsen2004,Lappin2005,McShane2016} and sluicing \cite{Anand2016} but none of these methods consider gapping constructions.

\section{Conclusion}

We presented two methods to recover elided predicates in sentences with gapping.
Our experiments suggest that both methods work equally well in a realistic end-to-end
setting. While in general, recall is still low, the oracle experiments
suggest that both methods can recover elided predicates from correct dependency trees,
which suggests that as parsers become more and more accurate, the gap recovery accuracy should 
also increase.

We also demonstrated that our method can be used to automatically add the enhanced UD representation
to UD treebanks in other languages than English. Apart from being useful in a parsing pipeline, we
therefore also expect our method to be useful for building enhanced UD treebanks.

\section*{Reproducibility}

All data, pre-trained models, system outputs as well as a package for running the enhancement procedure are available from \url{https://github.com/sebschu/naacl-gapping}.

\section*{Acknowledgments}
We thank the anonymous reviewers for their thoughtful feedback. Also thanks to Vera Gribanova and Boris Harizanov for continuous feedback throughout this project, and to Matthew Lamm for help with annotating the data.
This work was supported in part by gifts from Google, Inc. and IPSoft, Inc. The first author is also supported by a Goodan Family Graduate Fellowship.

\bibliography{references}

\begin{thebibliography}{}
\expandafter\ifx\csname natexlab\endcsname\relax\def\natexlab#1{#1}\fi

\bibitem[{Anand and Hardt(2016)}]{Anand2016}
Pranav Anand and Daniel Hardt. 2016.
\newblock \href{https://doi.org/10.18653/v1/D16-1131}{Antecedent selection for
  sluicing: {S}tructure and content}.
\newblock In {\em Proceedings of the 2016 Conference on Empirical Methods in
  Natural Language Processing (EMNLP 2016)\/}. pages 1234--1243.
\newblock \url{https://doi.org/10.18653/v1/D16-1131}.

\bibitem[{Bej{\v c}ek et~al.(2013)Bej{\v c}ek, Haji{\v c}ov{\'a}, Haji{\v c},
  J{\'i}nov{\'a}, Kettnerov{\'a}, Kol{\'a}{\v r}ov{\'a}, Mikulov{\'a},
  M{\'i}rovsk{\'y}, Nedoluzhko, Panevov{\'a}, Pol{\'a}kov{\'a}, {\v S}ev{\v
  c}{\'i}kov{\'a}, {\v S}t{\v e}p{\'a}nek, and Zik{\'a}nov{\'a}}]{PDT2013}
Eduard Bej{\v c}ek, Eva Haji{\v c}ov{\'a}, Jan Haji{\v c}, Pavl{\'i}na
  J{\'i}nov{\'a}, V{\'a}clava Kettnerov{\'a}, Veronika Kol{\'a}{\v r}ov{\'a},
  Marie Mikulov{\'a}, Ji{\v r}{\'i} M{\'i}rovsk{\'y}, Anna Nedoluzhko, Jarmila
  Panevov{\'a}, Lucie Pol{\'a}kov{\'a}, Magda {\v S}ev{\v c}{\'i}kov{\'a}, Jan
  {\v S}t{\v e}p{\'a}nek, and {\v S}{\'a}rka Zik{\'a}nov{\'a}. 2013.
\newblock \href{http://hdl.handle.net/11858/00-097C-0000-0023-1AAF-3}{{P}rague
  {D}ependency {T}reebank 3.0}.
\newblock {LINDAT}/{CLARIN} digital library at the Institute of Formal and
  Applied Linguistics, Charles University.
\newblock \url{http://hdl.handle.net/11858/00-097C-0000-0023-1AAF-3}.

\bibitem[{Bies et~al.(1995)Bies, Ferguson, Katz, MacIntyre, Tredinnick, Kim,
  Marcinkiewicz, and Schasberger}]{Bies1995}
Ann Bies, Mark Ferguson, Karen Katz, Robert MacIntyre, Victoria Tredinnick,
  Grace Kim, Mary~Ann Marcinkiewicz, and Britta Schasberger. 1995.
\newblock Bracketing guidelines for {T}reebank {II} style {P}enn {T}reebank
  project.
\newblock Technical report, University of Pennsylvania.

\bibitem[{Cai et~al.(2011)Cai, Chiang, and Goldberg}]{Cai2011}
Shu Cai, David Chiang, and Yoav Goldberg. 2011.
\newblock \href{http://www.aclweb.org/anthology/P11-2037}{Language-independent
  parsing with empty elements}.
\newblock In {\em Proceedings of the 49th Annual Meeting of the Association for
  Computational Linguistics: Human Language Technologies (ACL 2011)\/}. pages
  212--216.
\newblock \url{http://www.aclweb.org/anthology/P11-2037}.

\bibitem[{Campbell(2004)}]{Campbell2004}
Richard Campbell. 2004.
\newblock \href{https://doi.org/10.3115/1218955.1219037}{Using linguistic
  principles to recover empty categories}.
\newblock In {\em Proceedings of the 42nd Annual Meeting on Association for
  Computational Linguistics (ACL 2004)\/}.
\newblock \url{https://doi.org/10.3115/1218955.1219037}.

\bibitem[{Collins(1997)}]{Collins1997}
Michael Collins. 1997.
\newblock \href{https://doi.org/10.3115/976909.979620}{Three generative,
  lexicalised models for statistical parsing}.
\newblock In {\em Proceedings of the 35th Annual Meeting of the Association for
  Computational Linguistics (ACL 1997)\/}. pages 16--23.
\newblock \url{https://doi.org/10.3115/976909.979620}.

\bibitem[{Coppock(2001)}]{Coppock2001}
Elizabeth Coppock. 2001.
\newblock Gapping: {I}n defense of deletion.
\newblock In Mary Andronis, Christopher Ball, Heidi Elston, and Sylvain Neuvel,
  editors, {\em Papers from the 37th Meeting of the Chicago Linguistic
  Society\/}. Chicago Linguistic Society, Chicago, pages 133--148.

\bibitem[{Dienes and Dubey(2003{\natexlab{a}})}]{Dienes2003b}
P{\'{e}}ter Dienes and Amit Dubey. 2003{\natexlab{a}}.
\newblock \href{http://www.aclweb.org/anthology/W03-1005}{Antecedent recovery:
  {E}xperiments with a trace tagger}.
\newblock In {\em Proceedings of the 2003 Conference on Empirical Methods in
  Natural Language Processing (EMNLP 2003)\/}. pages 33--40.
\newblock \url{http://www.aclweb.org/anthology/W03-1005}.

\bibitem[{Dienes and Dubey(2003{\natexlab{b}})}]{Dienes2003}
P{\'{e}}ter Dienes and Amit Dubey. 2003{\natexlab{b}}.
\newblock \href{https://doi.org/10.3115/1075096.1075151}{Deep syntactic
  processing by combining shallow methods}.
\newblock In {\em Proceedings of the 41st Annual Meeting on Association for
  Computational Linguistics (ACL 2003)\/}. pages 431--438.
\newblock \url{https://doi.org/10.3115/1075096.1075151}.

\bibitem[{Dozat and Manning(2017)}]{Dozat2017}
Timothy Dozat and Christopher~D. Manning. 2017.
\newblock \href{https://openreview.net/pdf?id=Hk95PK9le}{Deep biaffine
  attention for neural dependency parsing}.
\newblock In {\em Proceedings of the 5th International Conference on Learning
  Representations (ICLR 2017)\/}. pages 1--8.
\newblock \url{https://openreview.net/pdf?id=Hk95PK9le}.

\bibitem[{Dozat et~al.(2017)Dozat, Qi, and Manning}]{Dozat2017b}
Timothy Dozat, Peng Qi, and Christopher~D. Manning. 2017.
\newblock \href{https://doi.org/10.18653/v1/K17-3002}{{S}tanford's graph-based
  neural dependency parser at the {CoNLL} 2017 {S}hared {T}ask}.
\newblock In {\em Proceedings of the CoNLL 2017 Shared Task: Multilingual
  Parsing from Raw Text to Universal Dependencies\/}. pages 20--30.
\newblock \url{https://doi.org/10.18653/v1/K17-3002}.

\bibitem[{Droganova and Zeman(2017)}]{Droganova2017}
Kira Droganova and Daniel Zeman. 2017.
\newblock \href{http://www.aclweb.org/anthology/W17-0406}{Elliptic
  constructions: {S}potting patterns in {UD} treebanks}.
\newblock In {\em Proceedings of the NoDaLiDa 2017 Workshop on Universal
  Dependencies (UDW 2017)\/}. pages 48--57.
\newblock \url{http://www.aclweb.org/anthology/W17-0406}.

\bibitem[{Dukes and Habash(2011)}]{Dukes2011}
Kais Dukes and Nizar Habash. 2011.
\newblock \href{http://www.aclweb.org/anthology/W11-2912}{One-step statistical
  parsing of hybrid dependency-constituency syntactic representations}.
\newblock In {\em Proceedings of the 12th International Conference on Parsing
  Technologies\/}. pages 92--103.
\newblock \url{http://www.aclweb.org/anthology/W11-2912}.

\bibitem[{Einarsson(1976)}]{Einarsson1976}
Jan Einarsson. 1976.
\newblock Talbankens skriftspr{\aa}kskonkordans.
\newblock Institutionen f\"or nordiska spr{\aa}k, {L}unds universitet.

\bibitem[{Farudi(2013)}]{Farudi2013}
Annahita Farudi. 2013.
\newblock {\em Gapping in {F}arsi: {A} Crosslinguistic Investigation\/}.
\newblock Ph.D. thesis, University of Massachusetts Amherst.
\newblock \url{http://scholarworks.umass.edu/dissertations/AAI3556244}.

\bibitem[{Ficler and Goldberg(2016)}]{Ficler2016}
Jessica Ficler and Yoav Goldberg. 2016.
\newblock \href{https://doi.org/10.18653/v1/P16-2012}{Improved parsing for
  argument-clusters coordination}.
\newblock In {\em Proceedings of the 54th Annual Meeting of the Association for
  Computational Linguistics (ACL 2016)\/}. pages 72--76.
\newblock \url{https://doi.org/10.18653/v1/P16-2012}.

\bibitem[{Gabbard et~al.(2006)Gabbard, Marcus, and Kulick}]{Gabbard2006}
Ryan Gabbard, Mitchell Marcus, and Seth Kulick. 2006.
\newblock \href{https://doi.org/10.3115/1220835.1220859}{Fully parsing the
  {P}enn {T}reebank}.
\newblock In {\em Proceedings of the Human Language Technology Conference of
  the North American Chapter of the ACL (NAACL 2006)\/}. pages 184--191.
\newblock \url{https://doi.org/10.3115/1220835.1220859}.

\bibitem[{Hardt(1997)}]{Hardt1997}
Daniel Hardt. 1997.
\newblock \href{http://www.aclweb.org/anthology/J97-4002}{An empirical approach
  to {VP} ellipsis}.
\newblock {\em Computational Linguistics\/} 23(4):525--541.
\newblock \url{http://www.aclweb.org/anthology/J97-4002}.

\bibitem[{Haugereid(2017)}]{Haugereid2017}
Petter Haugereid. 2017.
\newblock
  \href{http://cslipublications.stanford.edu/HPSG/2017/hpsg2017-haugereid.pdf}{An
  incremental approach to gapping and conjunction reduction}.
\newblock In {\em Proceedings of the 24th International Conference on
  Head-Driven Phrase Structure Grammar\/}. pages 179--198.
\newblock
  \url{http://cslipublications.stanford.edu/HPSG/2017/hpsg2017-haugereid.pdf}.

\bibitem[{Hayashi and Nagata(2016)}]{Hayashi2016}
Katsuhiko Hayashi and Masaaki Nagata. 2016.
\newblock \href{https://doi.org/10.18653/v1/P16-2016}{Empty element recovery by
  spinal parser operations}.
\newblock In {\em Proceedings of the 54th Annual Meeting of the Association for
  Computational Linguistics (ACL 2016)\/}. pages 95--100.
\newblock \url{https://doi.org/10.18653/v1/P16-2016}.

\bibitem[{Hockenmaier and Steedman(2007)}]{Hockenmaier2007}
Julia Hockenmaier and Mark Steedman. 2007.
\newblock \href{http://www.aclweb.org/anthology/J07-3004}{{CCGbank}: a corpus
  of {CCG} derivations and dependency structures extracted from the {P}enn
  {T}reebank}.
\newblock {\em Computational Linguistics\/} 33(3):355--396.
\newblock \url{http://www.aclweb.org/anthology/J07-3004}.

\bibitem[{Jackendoff(1971)}]{Jackendoff1971}
Ray~S. Jackendoff. 1971.
\newblock {Gapping and related rules}.
\newblock {\em Linguistic Inquiry\/} 2(1):21--35.

\bibitem[{Johnson(2014)}]{Johnson2014}
Kyle Johnson. 2014.
\newblock
  \href{http://people.umass.edu/kbj/homepage/Content/gapping.pdf}{Gapping}.
\newblock Unpublished manuscript, University of Massachusetts at Amherst.
\newblock \url{http://people.umass.edu/kbj/homepage/Content/gapping.pdf}.

\bibitem[{Johnson(2002)}]{Johnson2002}
Mark Johnson. 2002.
\newblock \href{https://doi.org/10.3115/1073083.1073107}{A simple
  pattern-matching algorithm for recovering empty nodes and their antecedents}.
\newblock In {\em Proceedings of the 40th Annual Meeting of the Association for
  Computational Linguistics (ACL 2002)\/}. pages 136--143.
\newblock \url{https://doi.org/10.3115/1073083.1073107}.

\bibitem[{Kato and Matsubara(2016)}]{Kato2016}
Yoshihide Kato and Shigeki Matsubara. 2016.
\newblock \href{https://doi.org/10.18653/v1/P16-1088}{Transition-based
  left-corner parsing for identifying {PTB}-style nonlocal dependencies}.
\newblock In {\em Proceedings of the 54th Annual Meeting of the Association for
  Computational Linguistics (ACL 2016)\/}. pages 930--940.
\newblock \url{https://doi.org/10.18653/v1/P16-1088}.

\bibitem[{Kay(1980)}]{Kay1980}
Martin Kay. 1980.
\newblock Algorithm schemata and data structures in syntactic processing.
\newblock Technical report, Xerox PARC.

\bibitem[{Klein and Manning(2001)}]{Klein2001}
Dan Klein and Christopher Manning. 2001.
\newblock \href{https://doi.org/10.3115/1073012.1073056}{Parsing with treebank
  grammars: empirical bounds, theoretical models, and the structure of the
  {P}enn {T}reebank}.
\newblock In {\em Proceedings of 39th Annual Meeting of the Association for
  Computational Linguistics (ACL 2001)\/}. pages 338--345.
\newblock \url{https://doi.org/10.3115/1073012.1073056}.

\bibitem[{Kubota and Levine(2016)}]{Kubota2016}
Yusuke Kubota and Robert Levine. 2016.
\newblock \href{https://doi.org/10.1007/s11049-015-9298-4}{Gapping as
  hypothetical reasoning}.
\newblock {\em Natural Language and Linguistic Theory\/} 34(1):107--156.
\newblock \url{https://doi.org/10.1007/s11049-015-9298-4}.

\bibitem[{Kummerfeld and Klein(2017)}]{Kummerfeld2017}
Jonathan~K. Kummerfeld and Dan Klein. 2017.
\newblock
  \href{https://www.transacl.org/ojs/index.php/tacl/article/view/1170}{Parsing
  with traces: {A}n {$O(n^4)$} algorithm and a structural representation}.
\newblock {\em Transactions of the Association for Computational Linguistics\/}
  5:441--454.
\newblock \url{https://www.transacl.org/ojs/index.php/tacl/article/view/1170}.

\bibitem[{Lappin(2005)}]{Lappin2005}
Shalom Lappin. 2005.
\newblock \href{https://doi.org/10.1075/cilt.263.03lap}{A sequenced model of
  anaphora and ellipsis resolution}.
\newblock In Ant{\'{o}}nio Branco, Tony McEnery, and Ruslan Mitkov, editors,
  {\em Anaphora Processing: Linguistic, Cognitive and Computational
  Modelling\/}, John Benjamins Publishing, Amsterdam, pages 3--16.
\newblock \url{https://doi.org/10.1075/cilt.263.03lap}.

\bibitem[{Levy and Manning(2004)}]{Levy2004}
Roger Levy and Christopher~D. Manning. 2004.
\newblock \href{https://doi.org/10.3115/1218955.1218997}{Deep dependencies from
  context-free statistical parsers}.
\newblock In {\em Proceedings of the 42nd Annual Meeting on Association for
  Computational Linguistics (ACL 2004)\/}. pages 327--334.
\newblock \url{https://doi.org/10.3115/1218955.1218997}.

\bibitem[{Marcus et~al.(1993)Marcus, Santorini, and Marcinkiewicz}]{Marcus1993}
Mitchell~P. Marcus, Beatrice Santorini, and Mary~Ann Marcinkiewicz. 1993.
\newblock \href{http://www.aclweb.org/anthology/J93-2004}{Building a large
  annotated corpus of {E}nglish: {T}he {P}enn {T}reebank}.
\newblock {\em Computational Linguistics\/} 19(2):313--330.
\newblock \url{http://www.aclweb.org/anthology/J93-2004}.

\bibitem[{McDonald et~al.(2005)McDonald, Pereira, Ribarov, and
  Haji{\v{c}}}]{McDonald2005}
Ryan McDonald, Fernando Pereira, Kiril Ribarov, and Jan Haji{\v{c}}. 2005.
\newblock \href{https://doi.org/10.3115/1220575.1220641}{Non-projective
  dependency parsing using spanning tree algorithms}.
\newblock In {\em Proceedings of Human Language Technology Conference and
  Conference on Empirical Methods in Natural Language Processing (HLT-EMNLP
  2005)\/}. pages 523--530.
\newblock \url{https://doi.org/10.3115/1220575.1220641}.

\bibitem[{McShane and Babkin(2016)}]{McShane2016}
Marjorie McShane and Petr Babkin. 2016.
\newblock Detection and resolution of verb phrase ellipsis.
\newblock {\em Linguistic Issues in Language Technology (LiLT)\/} 13(1):1--34.

\bibitem[{Mikolov et~al.(2013)Mikolov, Sutskever, Chen, Corrado, and
  Dean}]{Mikolov2013}
Tomas Mikolov, Ilya Sutskever, Kai Chen, Greg~S Corrado, and Jeff Dean. 2013.
\newblock Distributed representations of words and phrases and their
  compositionality.
\newblock In {\em Advances in Neural Information Processing Systems 26 (NIPS
  2013)\/}. pages 3111--3119.

\bibitem[{Needleman and Wunsch(1970)}]{Needleman1970}
Saul~B. Needleman and Christian~D. Wunsch. 1970.
\newblock \href{https://doi.org/10.1016/0022-2836(70)90057-4}{A general method
  applicable to the search for similarities in the amino acid sequence of two
  proteins}.
\newblock {\em Journal of Molecular Biology\/} 48(3):443--453.
\newblock \url{https://doi.org/10.1016/0022-2836(70)90057-4}.

\bibitem[{Nielsen(2004)}]{Nielsen2004}
Leif~Arda Nielsen. 2004.
\newblock \href{https://doi.org/10.3115/1220355.1220512}{Verb phrase ellipsis
  detection using automatically parsed text}.
\newblock In {\em Proceedings of the 20th International Conference on
  Computational Linguistics (COLING 2004)\/}. 1, pages 1093--1099.
\newblock \url{https://doi.org/10.3115/1220355.1220512}.

\bibitem[{Nivre et~al.(2017)Nivre, Agi{\'c}, Ahrenberg, Antonsen, Aranzabe,
  Asahara, Ateyah, Attia, Atutxa, Augustinus, Badmaeva, Ballesteros, Banerjee,
  Bank, Barbu~Mititelu, Bauer, Bengoetxea, Bhat, Bick, Bobicev, B{\"o}rstell,
  Bosco, Bouma, Bowman, Burchardt, Candito, Caron, Cebiro{\u g}lu~Eryi{\u g}it,
  Celano, Cetin, Chalub, Choi, Cinkov{\'a}, {\c C}{\"o}ltekin, Connor,
  Davidson, de~Marneffe, de~Paiva, Diaz~de Ilarraza, Dirix, Dobrovoljc, Dozat,
  Droganova, Dwivedi, Eli, Elkahky, Erjavec, Farkas, Fernandez~Alcalde, Foster,
  Freitas, Gajdo{\v s}ov{\'a}, Galbraith, Garcia, G{\"a}rdenfors, Gerdes,
  Ginter, Goenaga, Gojenola, G{\"o}k{\i}rmak, Goldberg, G{\'o}mez~Guinovart,
  Gonz{\'a}les~Saavedra, Grioni, Gr{\=u}z{\={\i}}tis, Guillaume, Habash,
  Haji{\v c}, Haji{\v c}~jr., H{\`a}~M{\~y}, Harris, Haug, Hladk{\'a},
  Hlav{\'a}{\v c}ov{\'a}, Hociung, Hohle, Ion, Irimia, Jel{\'{\i}}nek,
  Johannsen, J{\o}rgensen, Ka{\c s}{\i}kara, Kanayama, Kanerva, Kayadelen,
  Kettnerov{\'a}, Kirchner, Kotsyba, Krek, Laippala, Lambertino, Lando, Lee,
  L{\^e}~H{\`{\^o}}ng, Lenci, Lertpradit, Leung, Li, Li, Li, Ljube{\v s}i{\'c},
  Loginova, Lyashevskaya, Lynn, Macketanz, Makazhanov, Mandl, Manning, M{\u
  a}r{\u a}nduc, Mare{\v c}ek, Marheinecke, Mart{\'{\i}}nez~Alonso, Martins,
  Ma{\v s}ek, Matsumoto, {McDonald}, Mendon{\c c}a, Miekka, Missil{\"a},
  Mititelu, Miyao, Montemagni, More, Moreno~Romero, Mori, Moskalevskyi,
  Muischnek, M{\"u}{\"u}risep, Nainwani, Nedoluzhko, Ne{\v
  s}pore-B{\=e}rzkalne, Nguy{\~{\^e}}n~Th{\d i}, Nguy{\~{\^e}}n Th{\d i}~Minh,
  Nikolaev, Nurmi, Ojala, Osenova, {\"O}stling, {\O}vrelid, Pascual,
  Passarotti, Perez, Perrier, Petrov, Piitulainen, Pitler, Plank, Popel,
  Pretkalni{\c n}a, Prokopidis, Puolakainen, Pyysalo, Rademaker, Ramasamy,
  Rama, Ravishankar, Real, Reddy, Rehm, Rinaldi, Rituma, Romanenko, Rosa,
  Rovati, Sagot, Saleh, Samard{\v z}i{\'c}, Sanguinetti, Saul{\={\i}}te,
  Schuster, Seddah, Seeker, Seraji, Shen, Shimada, Sichinava, Silveira, Simi,
  Simionescu, Simk{\'o}, {\v S}imkov{\'a}, Simov, Smith, Stella, Straka,
  Strnadov{\'a}, Suhr, Sulubacak, Sz{\'a}nt{\'o}, Taji, Tanaka, Trosterud,
  Trukhina, Tsarfaty, Tyers, Uematsu, Ure{\v s}ov{\'a}, Uria, Uszkoreit,
  Vajjala, van Niekerk, van Noord, Varga, Villemonte de~la Clergerie, Vincze,
  Wallin, Washington, Wir{\'e}n, Wong, Yu, {\v Z}abokrtsk{\'y}, Zeldes, Zeman,
  and Zhu}]{Nivre2017a}
Joakim Nivre, {\v Z}eljko Agi{\'c}, Lars Ahrenberg, Lene Antonsen, Maria~Jesus
  Aranzabe, Masayuki Asahara, Luma Ateyah, Mohammed Attia, Aitziber Atutxa,
  Liesbeth Augustinus, Elena Badmaeva, Miguel Ballesteros, Esha Banerjee,
  Sebastian Bank, Verginica Barbu~Mititelu, John Bauer, Kepa Bengoetxea,
  Riyaz~Ahmad Bhat, Eckhard Bick, Victoria Bobicev, Carl B{\"o}rstell, Cristina
  Bosco, Gosse Bouma, Sam Bowman, Aljoscha Burchardt, Marie Candito, Gauthier
  Caron, G{\"u}l{\c s}en Cebiro{\u g}lu~Eryi{\u g}it, Giuseppe G.~A. Celano,
  Savas Cetin, Fabricio Chalub, Jinho Choi, Silvie Cinkov{\'a}, {\c C}a{\u
  g}r{\i} {\c C}{\"o}ltekin, Miriam Connor, Elizabeth Davidson, Marie-Catherine
  de~Marneffe, Valeria de~Paiva, Arantza Diaz~de Ilarraza, Peter Dirix, Kaja
  Dobrovoljc, Timothy Dozat, Kira Droganova, Puneet Dwivedi, Marhaba Eli, Ali
  Elkahky, Toma{\v z} Erjavec, Rich{\'a}rd Farkas, Hector Fernandez~Alcalde,
  Jennifer Foster, Cl{\'a}udia Freitas, Katar{\'{\i}}na Gajdo{\v s}ov{\'a},
  Daniel Galbraith, Marcos Garcia, Moa G{\"a}rdenfors, Kim Gerdes, Filip
  Ginter, Iakes Goenaga, Koldo Gojenola, Memduh G{\"o}k{\i}rmak, Yoav Goldberg,
  Xavier G{\'o}mez~Guinovart, Berta Gonz{\'a}les~Saavedra, Matias Grioni,
  Normunds Gr{\=u}z{\={\i}}tis, Bruno Guillaume, Nizar Habash, Jan Haji{\v c},
  Jan Haji{\v c}~jr., Linh H{\`a}~M{\~y}, Kim Harris, Dag Haug, Barbora
  Hladk{\'a}, Jaroslava Hlav{\'a}{\v c}ov{\'a}, Florinel Hociung, Petter Hohle,
  Radu Ion, Elena Irimia, Tom{\'a}{\v s} Jel{\'{\i}}nek, Anders Johannsen,
  Fredrik J{\o}rgensen, H{\"u}ner Ka{\c s}{\i}kara, Hiroshi Kanayama, Jenna
  Kanerva, Tolga Kayadelen, V{\'a}clava Kettnerov{\'a}, Jesse Kirchner, Natalia
  Kotsyba, Simon Krek, Veronika Laippala, Lorenzo Lambertino, Tatiana Lando,
  John Lee, Ph{\uhorn}{\ohorn}ng L{\^e}~H{\`{\^o}}ng, Alessandro Lenci, Saran
  Lertpradit, Herman Leung, Cheuk~Ying Li, Josie Li, Keying Li, Nikola Ljube{\v
  s}i{\'c}, Olga Loginova, Olga Lyashevskaya, Teresa Lynn, Vivien Macketanz,
  Aibek Makazhanov, Michael Mandl, Christopher~D. Manning, C{\u a}t{\u a}lina
  M{\u a}r{\u a}nduc, David Mare{\v c}ek, Katrin Marheinecke, H{\'e}ctor
  Mart{\'{\i}}nez~Alonso, Andr{\'e} Martins, Jan Ma{\v s}ek, Yuji Matsumoto,
  Ryan {McDonald}, Gustavo Mendon{\c c}a, Niko Miekka, Anna Missil{\"a}, C{\u
  a}t{\u a}lin Mititelu, Yusuke Miyao, Simonetta Montemagni, Amir More, Laura
  Moreno~Romero, Shinsuke Mori, Bohdan Moskalevskyi, Kadri Muischnek, Kaili
  M{\"u}{\"u}risep, Pinkey Nainwani, Anna Nedoluzhko, Gunta Ne{\v
  s}pore-B{\=e}rzkalne, L{\uhorn}{\ohorn}ng Nguy{\~{\^e}}n~Th{\d i},
  Huy{\`{\^e}}n Nguy{\~{\^e}}n Th{\d i}~Minh, Vitaly Nikolaev, Hanna Nurmi,
  Stina Ojala, Petya Osenova, Robert {\"O}stling, Lilja {\O}vrelid, Elena
  Pascual, Marco Passarotti, Cenel-Augusto Perez, Guy Perrier, Slav Petrov,
  Jussi Piitulainen, Emily Pitler, Barbara Plank, Martin Popel, Lauma
  Pretkalni{\c n}a, Prokopis Prokopidis, Tiina Puolakainen, Sampo Pyysalo,
  Alexandre Rademaker, Loganathan Ramasamy, Taraka Rama, Vinit Ravishankar,
  Livy Real, Siva Reddy, Georg Rehm, Larissa Rinaldi, Laura Rituma, Mykhailo
  Romanenko, Rudolf Rosa, Davide Rovati, Beno{\^{\i}}t Sagot, Shadi Saleh,
  Tanja Samard{\v z}i{\'c}, Manuela Sanguinetti, Baiba Saul{\={\i}}te,
  Sebastian Schuster, Djam{\'e} Seddah, Wolfgang Seeker, Mojgan Seraji,
  Mo~Shen, Atsuko Shimada, Dmitry Sichinava, Natalia Silveira, Maria Simi, Radu
  Simionescu, Katalin Simk{\'o}, M{\'a}ria {\v S}imkov{\'a}, Kiril Simov, Aaron
  Smith, Antonio Stella, Milan Straka, Jana Strnadov{\'a}, Alane Suhr, Umut
  Sulubacak, Zsolt Sz{\'a}nt{\'o}, Dima Taji, Takaaki Tanaka, Trond Trosterud,
  Anna Trukhina, Reut Tsarfaty, Francis Tyers, Sumire Uematsu, Zde{\v n}ka
  Ure{\v s}ov{\'a}, Larraitz Uria, Hans Uszkoreit, Sowmya Vajjala, Daniel van
  Niekerk, Gertjan van Noord, Viktor Varga, Eric Villemonte de~la Clergerie,
  Veronika Vincze, Lars Wallin, Jonathan~North Washington, Mats Wir{\'e}n,
  Tak-sum Wong, Zhuoran Yu, Zden{\v e}k {\v Z}abokrtsk{\'y}, Amir Zeldes,
  Daniel Zeman, and Hanzhi Zhu. 2017.
\newblock \href{http://hdl.handle.net/11234/1-2515}{{U}niversal {D}ependencies
  2.1}.
\newblock {LINDAT}/{CLARIN} digital library at the Institute of Formal and
  Applied Linguistics ({{\'U}FAL}), Faculty of Mathematics and Physics, Charles
  University.
\newblock \url{http://hdl.handle.net/11234/1-2515}.

\bibitem[{Nivre et~al.(2016)Nivre, de~Marneffe, Ginter, Goldberg, Haji{\v{c}},
  Manning, McDonald, Petrov, Pyysalo, Silveira, Tsarfaty, and
  Zeman}]{Nivre2016}
Joakim Nivre, Marie-Catherine de~Marneffe, Filip Ginter, Yoav Goldberg, Jan
  Haji{\v{c}}, Christopher~D. Manning, Ryan McDonald, Slav Petrov, Sampo
  Pyysalo, Natalia Silveira, Reut Tsarfaty, and Daniel Zeman. 2016.
\newblock
  \href{http://www.lrec-conf.org/proceedings/lrec2016/pdf/348_Paper.pdf}{{U}niversal
  {D}ependencies v1: {A} multilingual treebank collection}.
\newblock In {\em Proceedings of the Tenth International Conference on Language
  Resources and Evaluation (LREC 2016)\/}. pages 1659--1666.
\newblock
  \url{http://www.lrec-conf.org/proceedings/lrec2016/pdf/348_Paper.pdf}.

\bibitem[{Noreen(1989)}]{Noreen1989}
Eric~W. Noreen. 1989.
\newblock {\em {Computer-Intensive Methods for Testing Hypotheses}\/}.
\newblock John Wiley {\&} Sons, New York, NY.

\bibitem[{Oepen et~al.(2015)Oepen, Kuhlmann, Miyao, Zeman, Cinkova, Flickinger,
  Hajic, and Uresova}]{Oepen2015}
Stephan Oepen, Marco Kuhlmann, Yusuke Miyao, Daniel Zeman, Silvie Cinkova, Dan
  Flickinger, Jan Hajic, and Zdenka Uresova. 2015.
\newblock \href{https://doi.org/10.18653/v1/S15-2153}{{SemEval} 2015 {T}ask 18:
  {B}road-coverage semantic dependency parsing}.
\newblock In {\em Proceedings of the 9th International Workshop on Semantic
  Evaluation (SemEval 2015)\/}. pages 915--926.
\newblock \url{https://doi.org/10.18653/v1/S15-2153}.

\bibitem[{Ohta et~al.(2002)Ohta, Tateisi, and Kim}]{Ohta2002}
Tomoko Ohta, Yuka Tateisi, and Jin-Dong Kim. 2002.
\newblock \href{https://doi.org/10.3115/1289189.1289260}{The {GENIA} corpus:
  {A}n annotated research abstract corpus in molecular biology domain}.
\newblock {\em Proceedings of the Second International Conference on Human
  Language Technology Research (HLT 2002)\/} pages 82--86.
\newblock \url{https://doi.org/10.3115/1289189.1289260}.

\bibitem[{Osborne(2006)}]{Osborne2006}
Timothy Osborne. 2006.
\newblock {Gapping vs. non-gapping coordination}.
\newblock {\em Linguistische Berichte\/} 207:307--337.

\bibitem[{Pad{\'{o}}(2006)}]{Pado2006}
Sebastian Pad{\'{o}}. 2006.
\newblock {\em User's guide to \texttt{sigf}: {S}ignificance testing by
  approximate randomisation\/}.
\newblock \url{https://nlpado.de/~sebastian/software/sigf.shtml}.

\bibitem[{Pennington et~al.(2014)Pennington, Socher, and
  Manning}]{Pennington2014}
Jeffrey Pennington, Richard Socher, and Christopher Manning. 2014.
\newblock \href{https://doi.org/10.3115/v1/D14-1162}{{Glove}: {G}lobal vectors
  for word representation}.
\newblock In {\em Proceedings of the 2014 Conference on Empirical Methods in
  Natural Language Processing (EMNLP 2014)\/}. pages 1532--1543.
\newblock \url{https://doi.org/10.3115/v1/D14-1162}.

\bibitem[{Ross(1970)}]{Ross1970}
John~Robert Ross. 1970.
\newblock Gapping and the order of constituents.
\newblock In Manfred Bierwisch and Karl~Erich Heidolph, editors, {\em Progress
  in Linguistics\/}, De Gruyter, The Hague, pages 249--259.

\bibitem[{Schmid(2006)}]{Schmid2006}
Helmut Schmid. 2006.
\newblock \href{https://doi.org/10.3115/1220175.1220198}{Trace prediction and
  recovery with unlexicalized {PCFG}s and slash features}.
\newblock In {\em Proceedings of the 21st International Conference on
  Computational Linguistics and 44th Annual Meeting of the ACL (ACL 2006)\/}.
  pages 177--184.
\newblock \url{https://doi.org/10.3115/1220175.1220198}.

\bibitem[{Schuster et~al.(2017)Schuster, Lamm, and Manning}]{Schuster2017}
Sebastian Schuster, Matthew Lamm, and Christopher~D. Manning. 2017.
\newblock \href{http://www.aclweb.org/anthology/W17-0416}{Gapping constructions
  in {U}niversal {D}ependencies v2}.
\newblock In {\em Proceedings of the NoDaLiDa 2017 Workshop on Universal
  Dependencies (UDW 2017)\/}. pages 123--132.
\newblock \url{http://www.aclweb.org/anthology/W17-0416}.

\bibitem[{Schuster and Manning(2016)}]{Schuster2016}
Sebastian Schuster and Christopher~D. Manning. 2016.
\newblock
  \href{http://www.lrec-conf.org/proceedings/lrec2016/pdf/779_Paper.pdf}{{E}nhanced
  {E}nglish {U}niversal {D}ependencies: {A}n improved representation for
  natural language understanding tasks}.
\newblock In {\em Proceedings of the Tenth International Conference on Language
  Resources and Evaluation (LREC 2016)\/}. pages 2371--2378.
\newblock
  \url{http://www.lrec-conf.org/proceedings/lrec2016/pdf/779_Paper.pdf}.

\bibitem[{Seeker et~al.(2012)Seeker, Farkas, Bohnet, Schmid, and
  Kuhn}]{Seeker2012}
Wolfgang Seeker, Rich{\'{a}}rd Farkas, Bernd Bohnet, Helmut Schmid, and Jonas
  Kuhn. 2012.
\newblock \href{http://www.aclweb.org/anthology/C12-2105}{Data-driven
  dependency parsing with empty heads}.
\newblock In {\em Proceedings of COLING 2012\/}. pages 1081--1090.
\newblock \url{http://www.aclweb.org/anthology/C12-2105}.

\bibitem[{Silveira et~al.(2014)Silveira, Dozat, {de Marneffe}, Bowman, Connor,
  Bauer, and Manning}]{Silveira2014}
Natalia Silveira, Timothy Dozat, Marie-Catherine {de Marneffe}, Samuel Bowman,
  Miriam Connor, John Bauer, and Christopher~D Manning. 2014.
\newblock
  \href{http://www.lrec-conf.org/proceedings/lrec2014/pdf/1089_Paper.pdf}{A
  gold standard dependency corpus for {E}nglish}.
\newblock In {\em Proceedings of the Ninth International Conference on Language
  Resources and Evaluation (LREC 2014)\/}. pages 2897--2904.
\newblock
  \url{http://www.lrec-conf.org/proceedings/lrec2014/pdf/1089_Paper.pdf}.

\bibitem[{Simk{\'{o}} and Vincze(2017)}]{Simko2017}
Katalin~Ilona Simk{\'{o}} and Veronika Vincze. 2017.
\newblock \href{http://www.aclweb.org/anthology/W17-6527}{{H}ungarian copula
  constructions in dependency syntax and parsing}.
\newblock In {\em Proceedings of the Fourth International Conference on
  Dependency Linguistics (Depling 2017)\/}. pages 240--247.
\newblock \url{http://www.aclweb.org/anthology/W17-6527}.

\bibitem[{Steedman(1990)}]{Steedman1990}
Mark Steedman. 1990.
\newblock \href{https://doi.org/10.1007/BF00630734}{Gapping as constituent
  coordination}.
\newblock {\em Linguistics and Philosophy\/} 13(2):207--263.
\newblock \url{https://doi.org/10.1007/BF00630734}.

\bibitem[{Toosarvandani(2016)}]{Toosarvandani2016}
Maziar Toosarvandani. 2016.
\newblock \href{https://doi.org/10.1162/LING_a_00216}{Embedding the antecedent
  in gapping: {L}ow coordination and the role of parallelism}.
\newblock {\em Linguistic Inquiry\/} 47(2):381--390.
\newblock \url{https://doi.org/10.1162/LING_a_00216}.

\bibitem[{Woods(1970)}]{Woods1970}
William~A. Woods. 1970.
\newblock Transition network grammars for natural language analysis.
\newblock {\em Communications of the ACM\/} 13(10):591--606.

\bibitem[{Woods(1973)}]{Woods1973}
William~A. Woods. 1973.
\newblock An experimental parsing system for transition network grammars.
\newblock In Randall Rustin, editor, {\em Natural Language Processing\/},
  Algorithmics Press, New York, pages 111--154.

\bibitem[{Wyngaerd(2007)}]{Wyngaerd2007}
G.~Vanden Wyngaerd. 2007.
\newblock
  \href{https://lirias.kuleuven.be/bitstream/123456789/408979/1/09HRPL&L02.pdf}{Gapping
  constituents}.
\newblock Unpublished manuscript, FWO/K.U. Brussel.
\newblock
  \url{https://lirias.kuleuven.be/bitstream/123456789/408979/1/09HRPL&L02.pdf}.

\bibitem[{Yeh(2000)}]{Yeh2000}
Alexander Yeh. 2000.
\newblock \href{http://www.aclweb.org/anthology/C00-2137}{More accurate tests
  for the statistical significance of result differences}.
\newblock In {\em Proceedings of the The 18th International Conference on
  Computational Linguistics (COLING 2000)\/}. pages 947--953.
\newblock \url{http://www.aclweb.org/anthology/C00-2137}.

\bibitem[{Zeman et~al.(2017)Zeman, Popel, Straka, Haji{\v{c}}, Nivre, Ginter,
  Luotolahti, Pyysalo, Petrov, Potthast, Tyers, Badmaeva, G{\"{o}}krmak,
  Nedoluzhko, Cinkov{\'{a}}, Haji{\v{c}}~jr., Hlav{\'{a}}{\v{c}}ov{\'{a}},
  Kettnerov{\'{a}}, Ure{\v{s}}ov{\'{a}}, Kanerva, Ojala, Missil{\"{a}},
  Manning, Schuster, Reddy, Taji, Habash, Leung, {de Marneffe}, Sanguinetti,
  Simi, Kanayama, de~Paiva, Droganova, {Mart{\'{i}}nez Alonso}, Uszkoreit,
  Macketanz, Burchardt, Harris, Marheinecke, Rehm, Kayadelen, Attia, Elkahky,
  Yu, Pitler, Lertpradit, Mandl, Kirchner, {Fernandez Alcalde}, Strnadova,
  Banerjee, Manurung, Stella, Shimada, Kwak, Mendon{\c{c}}a, Lando, Nitisaroj,
  and Li}]{Zeman2017}
Daniel Zeman, Martin Popel, Milan Straka, Jan Haji{\v{c}}, Joakim Nivre, Filip
  Ginter, Juhani Luotolahti, Sampo Pyysalo, Slav Petrov, Martin Potthast,
  Francis Tyers, Elena Badmaeva, Memduh G{\"{o}}krmak, Anna Nedoluzhko, Silvie
  Cinkov{\'{a}}, Jan Haji{\v{c}}~jr., Jaroslava Hlav{\'{a}}{\v{c}}ov{\'{a}},
  V{\'{a}}clava Kettnerov{\'{a}}, Zdenka Ure{\v{s}}ov{\'{a}}, Jenna Kanerva,
  Stina Ojala, Anna Missil{\"{a}}, Christopher~D. Manning, Sebastian Schuster,
  Siva Reddy, Dima Taji, Nizar Habash, Herman Leung, Marie-Catherine {de
  Marneffe}, Manuela Sanguinetti, Maria Simi, Hiroshi Kanayama, Valeria
  de~Paiva, Kira Droganova, H{\v{e}}ctor {Mart{\'{i}}nez Alonso}, Hans
  Uszkoreit, Vivien Macketanz, Aljoscha Burchardt, Kim Harris, Katrin
  Marheinecke, Georg Rehm, Tolga Kayadelen, Mohammed Attia, Ali Elkahky,
  Zhuoran Yu, Emily Pitler, Saran Lertpradit, Michael Mandl, Jesse Kirchner,
  Hector {Fernandez Alcalde}, Jana Strnadova, Esha Banerjee, Ruli Manurung,
  Antonio Stella, Atsuko Shimada, Sookyoung Kwak, Gustavo Mendon{\c{c}}a,
  Tatiana Lando, Rattima Nitisaroj, and Josie Li. 2017.
\newblock \href{https://doi.org/10.18653/v1/K17-3001}{{CoNLL} 2017 {S}hared
  {T}ask: {M}ultilingual parsing from raw text to {U}niversal {D}ependencies}.
\newblock In {\em Proceedings of the CoNLL 2017 Shared Task: Multilingual
  Parsing from Raw Text to Universal Dependencies\/}. pages 1--19.
\newblock \url{https://doi.org/10.18653/v1/K17-3001}.

\end{thebibliography}
\bibliographystyle{acl_natbib}

\end{document}